\documentclass[conference]{IEEEtran}

\usepackage{cite}
\usepackage{pgf}
\usepackage{multirow}
\usepackage{graphicx}
\usepackage{amsmath}

\begin{document}

\title{An Experimental Study of Class Imbalance in Federated Learning}

\author{\IEEEauthorblockN{1\textsuperscript{st} Chenguang Xiao}
\IEEEauthorblockA{\textit{School of Computer Science}\\
\textit{University of Birmingham}\\
Edgbaston, Birmingham B15 2TT\\
Email: cxx075@student.bham.ac.uk}
\and
\IEEEauthorblockN{2\textsuperscript{nd} Shuo Wang}
\IEEEauthorblockA{\textit{School of Computer Science}\\
\textit{University of Birmingham}\\
Edgbaston, Birmingham B15 2TT\\
Email: S.Wang.2@bham.ac.uk}
}

\maketitle

\begin{abstract}
Federated learning is a distributed machine learning paradigm that trains a global model for prediction based on several local models at clients while local data privacy is preserved.
Class imbalance is believed to be one of the factors that degrades the global model performance.
However, there has been very little research on if and how class imbalance can affect the global performance in various imbalance scenarios.
Class imbalance in federated learning is much more complex than that in traditional non-distributed machine learning, due to different class imbalance situations at local clients.
Class imbalance needs to be re-defined in distributed learning environments, so that corresponding solutions can be proposed.
In this paper, first, we propose two new metrics to define class imbalance -- the global class imbalance degree (MID) and the local difference of class imbalance among clients (WCS).
Class imbalance is categorized into four scenarios under the definition.
Then, we conduct extensive experiments to analyze the impact of class imbalance on the global performance in various scenarios.
Our results show that a higher MID and a larger WCS degrade more the performance of the global model.
Besides, WCS is shown to slow down the convergence of the global model by misdirecting the optimization.
\end{abstract}


\section{Introduction}
As the rapid development of advanced computing hardware and machine learning algorithms, edge computing and ubiquitous computing systems have sprung forth. Local devices, such as mobile phones and wearable devices, have become a major source of data~\cite{Smith2017, Li2020, li2019survey}. A large number of devices are interconnected and are equipped with sensors that constantly generate potentially useful data~\cite{BrendanMcMahan2017}. To learn from such data, local data from clients have to be gathered together. However, these local generated data tend to contain sensitive information, such as end users' personal information and clients' medical records. Transmitting data among devices directly can cause privacy leakage and security issues.
Traditional machine learning approaches that collect and centralize user data will become legally impossible. Federated learning was thus proposed to learn from data, protect data privacy and improve network security~\cite{BrendanMcMahan2017}.

Federated learning trains a global model at the central server based on a group of local models trained and maintained at the clients~\cite{kairouz2019advances, konevcny2016federated}. Instead of transmitting data to the central server, only intermediate local model updates are communicated periodically with the server. In each round of training, the central server selects a group of clients and broadcasts the current global model to them. Then, the selected clients train the received model using their local data and feedback the model updates to the server. Lastly, the central server aggregates the updates from the clients. This iterative training process continues across the network~\cite{konevcny2016federated}.

Federated learning is able to make use of local data for training without violating privacy or breaking data island between clients. Currently, federated learning has been deployed by major service providers and plays a critical role in privacy-sensitive applications~\cite{Li2020}.
While most research in federated learning focuses on reducing communication loads and protecting data privacy~\cite{Li2020, kairouz2019advances, Smith2017}, little work has looked into how local data can affect the performance of the global model.
Data from different clients are always not independently and identically distributed (Non-IID). Class imbalance is a type of non-iid distributions in federated learning environments. In a classification task, such as fraudulent phone call detection and diagnoses of rare diseases, class imbalance refers to the situation where some classes of data (minority) are significantly under-represented compared to other classes (majority). In traditional non-distributed machine learning, class imbalance can cause great performance degradation, especially the poor accuracy on minority classes~\cite{japkowicz2002class, ali2013classification}. 

Class imbalance is also common in federated learning.
For example, in real-world health data, severe class imbalance is a norm, rather than an exception~\cite{choudhury2019predicting}.
However, it is unclear if and how local class imbalance in federated learning can affect the performance of the global model.
Compared with the traditional non-distributed learning case, class imbalance in federated learning is much more complex.
Local models and their regular updates could be affected by local imbalanced data, but it is unclear how many affected clients or how the severity of local imbalance may cause a global degradation.
Meanwhile, the class imbalance status between clients can vary.
For example, a client A has a minority class c1 and a majority class c2, but a client B has class c1 as majority and class c2 as minority.
Therefore, various class imbalance in federated learning should be categorised properly for potential solution to tackle them.
This paper aims at providing a full understanding of the impact of class imbalance in federated learning that will shed lights on suitable solutions of tackling class imbalance in federated learning.
We will answer three specific questions in this paper:
\begin{enumerate}
    \item How should we define class imbalance in federated learning? This will include a global class imbalance degree (applicable to multi-class data) and the differences of class imbalance between clients. 
    \item Would the local and global class imbalance affect the performance of the global model and how?
    \item Would the class imbalance difference between clients affect the performance of the global model and how?
\end{enumerate}

The contribution of this paper are listed bellow:
\begin{itemize}
    \item We propose two new metrics to measure class imbalance in federated learning: Global Imbalance Degree using Multiclass Imbalance Degree (MID) and Local and Global Imbalance Relation using Weighted Cosine Similarity (WCS). 
    \item Based on the new definition, we conduct extensive experiments on real-world datasets to investigate the impact of class imbalance. Four different scenarios are considered. Results show that the global class imbalance degrades the global model performance. The difference of local class imbalance also causes global performance degradation and slows down the model convergence.
\end{itemize}

The rest of this paper are organized as below. Section~\ref{sec:rw} presents related works. We define class imbalance in federated learning in Section~\ref{sec:cid}. Section~\ref{sec:cis} presents 4 class imbalance scenarios to be investigated and describes the datasets used in our experiments. Section~\ref{sec:exp} provides the experimental analysis. We conclude this paper and discuss the possible future work in Section~\ref{sec:con}.

\section{Related Works}
\label{sec:rw}

There is extensive investigation into class imbalance in non-distributed machine learning~\cite{Guo2008, Ali2015}. The impact of class imbalance depends on the imbalance level, concept complexity and size of training data~\cite{japkowicz2002class}. Lack of information caused by small sample size, class overlapping, and small disjuncts within class are main reasons of class imbalance causing performance degradation~\cite{Ali2015}. To tackle different types of class imbalance, the traditional approaches can be classified into five groups -- sampling approaches, re-weighting approaches, feature selection, one class learning, cost-sensitive learning~\cite{ling2008cost} and ensemble learning~\cite{Liu2009}. 

The nature of federated learning makes it different from non-distributed machine learning when dealing with class imbalance. In federated leaning, class imbalance presented at local data may or may not result in global class imbalance when the central server aggregates the model updates. Therefore, we need to separately define and discuss local class imbalance and global class imbalance. Furthermore, the local and global class imbalance can be totally different~\cite{Wang2020}. A majority class at some clients can be the minority class at the global level, and vice versa. The privacy protection of federated learning further increases the difficulty of estimating the class imbalance degree at the central server. As a result, existing class imbalance mitigating approaches can be only used locally, which may not help with global performance.

A few very recent papers~\cite{Wang2020, Duan2021, Sarkar2020, yang2020federated} have noticed the negative impact of class imbalance in federated learning and proposed techniques to tackle it. 
\textbf{Fed-Focal Loss}~\cite{Sarkar2020} used a modified loss function that down-weights the loss of well-classified samples based on Binary Cross Entropy (BCE) Loss. By doing so, the majority class with a larger number of examples contributes less to the model when it reaches high prediction accuracy. Correspondingly, the minority-class samples contribute more to the local model.
\textbf{Astraea}~\cite{Duan2021} added mediators between the central server and clients to re-balance the datasets. Imbalanced clients are rearranged to different balance mediators according to their imbalance levels and label distributions. Within the mediator, the clients perform training sequentially on a single balanced dataset. Then, the mediators communicate with the central server in parallel as clients in original federated learning.
\textbf{Ratio Loss}~\cite{Wang2020} employed a monitor scheme on the server to estimate local class imbalance without asking for label distributions. The monitoring scheme uses the relation between the gradient magnitude and the sample quantity to estimate global class imbalance at the server. Then, Ratio Loss based on BCE Loss is deployed at the local training process to strengthen the impact of minority-class examples.
Similarly, Yang et al.~\cite{yang2020federated} proposed a local class imbalance estimator based on gradient magnitude. Then clients selection is used to achieve class balance globally.

The aforementioned papers proposed new techniques to tackle class imbalance in specific setting, which shows the necessity of studying the class imbalance issue in federated learning.
However, they all treated a small part of class imbalance scenario as the full picture of class imbalance in federated learning.
Astraea was only valid with slightly global imbalance, while Ratio Loss can do nothing with a balanced global dataset consisted of imbalance local datasets.
Ratio Loss addressed the impact of mismatch between local and global imbalance, while the conclusion is drawn based on particular experiments setting without considering more general case with other imbalance degree.
In addition, Metrics used in those works include imbalance ratio, cosine similarity cannot fully reflect the imbalance states. Likelihood-ratio imbalanced degree (LRID)~\cite{Zhu2018} failed to measure the multiclass imbalance degree as well. The comparison between those metrics and ours will be discussed in Section~\ref{sec:cid}.
This paper thus aims at an in-depth understanding of class imbalance in various federated learning scenarios, which will help to develop the most suitable solutions in the future.

\section{Class Imbalance Definition}
\label{sec:cid}

Local imbalance and global imbalance were briefly mentioned in ~\cite{Wang2020} as two types of class imbalance in federated learning.
Their experiment on dedicated setting showed that a mismatch between local and global imbalance leads to global model performance degradation.
However, there is no clear definition to measure the global class imbalance degree and the relation between local and global imbalance. In this section, we propose two metrics to define class imbalance status in federated learning environments. 

\subsection{Federated Learning Problem Formulation}

Assume there are $P$ clients with local dataset $D_1,\dots,D_P$ in a size of $n_1, \dots, n_P$ respectively. If merging them together, the global dataset $D$ has $C$ classes and $N$ samples in total. At global time $t$, the global model is denoted as $w^t$. The selected client $p$ performs local training to derive a new local model $w_p^{t+1}$ by:
\begin{equation}
w_p^{t+1} = w_p^t - \nabla L(w^t, D_p)
\end{equation}
where $L(w^t, D_p)$ denotes the loss of model $w^t$ on dataset $D_p$. The update global model following the FedAvg~\cite{BrendanMcMahan2017} will be:
\begin{equation}
    w^{t+1} = w^t - \sum_{i=1}^P \frac{n_i}{N} \nabla L(w^t, D_i)
\end{equation}

\subsection{Global Imbalance Degree.}
To measure the global imbalance degree, we are inspired by  (LRID)~\cite{Zhu2018}. It was designed to measure class imbalance level in multi-class datasets. The commonly used Imbalance Ratio (denoted as $\Gamma$ below), referred to as the ratio between the numbers of the majority and minority classes, cannot fully describe the imbalance status in multi-class data as only two classes are considered. For a dataset with $N$ data samples and $C$ possible classes, the number of samples with label $c$ is $n_c$. The class imbalance level according to~\cite{Zhu2018} is defined as:
\begin{equation}
    LRID=-2\sum_{c=1}^C n_c\ln \frac{N}{Cn_c}
    \label{equ:lrid}
\end{equation}
The $LRID$ of an absolutely class balanced dataset is $0$. The larger $LRID$ is, the more class imbalanced the dataset is. However, $LRID$ is sensitive to the size of datasets. For two datasets $D_1$ and $D_2$ where $D_2$ contains exactly $k$ times the samples of $D_1$ for each class,  
$LRID_2$ based on Equation~\eqref{equ:lrid} becomes $-2k\sum_{c=1}^C n_c\ln \frac{N}{Cn_c}$, which is $k$ times larger than that of $D_1$ as Equation~\eqref{equ:lrid} even though the proportion of each class remains the same.

Given an extreme case where a $C$-class dataset contains $\left[N, 0, \dots, 0\right]$ samples for each class, $LRID_{extreme}=2N\log C$ according to Equation~\eqref{equ:lrid}. This $LRID$ value changes with the total sample number $N$, which is misleading as a measure of class imbalance. Therefore we improve LRID and propose Multiclass Imbalance Degree (MID): 

\begin{equation}
    MID =\frac{LRID}{LRID_{extreme}} = \sum_{c=1}^C \frac{n_c}{N}\log_C \frac{Cn_c}{N}
\end{equation}

MID eliminates the impact of the size of dataset and ranges between 0 and 1. MID equal to 0 implies a strictly balanced dataset. The larger the MID, the more imbalanced the dataset is. 
In our experiments, we use $MID$ to express how class imbalanced the global dataset $D$ is. 

\subsection{Local and Global Imbalance Relation.}
Mean cosine similarity (MCS) has been used to evaluate the mismatch between local and global imbalance~\cite{Wang2020}.
Cosine similarity of vector $A$ and $B$ is defined as
\begin{equation}
    similarity(A, B) = cos(\theta) = \frac{A \cdot B}{\parallel A \parallel_2 \parallel B \parallel_2}
\end{equation}
where $\theta$ is the angle between $A$ and $B$ and $\parallel A \parallel$ denotes the L2 norm of vector $A$. Label distribution vector of client $j$ is $l_j=\left[n_j^1,\dots,n_j^c, \dots, n_j^C\right]$ where $n_j^c$ is the number of samples with label $c$. The Global label distribution vector is $L=\left[\sum_{i=1}^P n_i^1, \dots, \sum_{i=1}^P n_i^C\right]$. Mean cosine similarity averages the similarity of global label distribution vector $L$ and local label distribution vector $l$ as below:
\begin{equation}
    MCS = \frac{1}{P}\sum_{i=1}^P \frac{L \cdot l_i}{\parallel L \parallel_2 \parallel l_i \parallel_2}
    \label{equ:mcs}
\end{equation}
It treats all clients equally and does not consider the sample size, which can be misleading. For example, a two-client federated dataset with label distribution vectors $l_1=[100, 99]$ (client1) and $l_2=[0,1]$ (client2) has $MCS=1/2(\frac{L \cdot l_1}{\parallel L \parallel_2\parallel l_1 \parallel_2} + \frac{L \cdot l_2}{\parallel L \parallel_2\parallel l_2 \parallel_2})=0.853$ according to Equation~\eqref{equ:mcs}. However, the similarity of global and local class imbalance should be nearly 1 as client2 contributes little to the global model. In this case, the small local dataset with extreme class imbalance leads to a biased estimation of the local and global imbalance relation when using MCS.
Therefore we propose Weighted Cosine Similarity (WCS) to measure the relationship between local and global imbalance that considers the contribution of local datasets.
WCS is defined as:

\begin{equation} \label{equ:wcs}
\begin{aligned}
WCS &=\sum_{i=1}^P \frac{\parallel l_i \parallel_1}{\parallel L \parallel_1} similarity(L,l_i) \\
 &=\sum_{i=1}^P \frac{\parallel l_i \parallel_1 L \cdot l_i}{\parallel L \parallel_1 \parallel L \parallel_2 \parallel l_i \parallel_2} \\
 &=\frac{1}{\parallel L \parallel_1 \parallel L \parallel_2} \sum_{i=1}^P \frac{\parallel l_i \parallel_1}{\parallel l_i \parallel_2}  L \cdot l_i
\end{aligned}
\end{equation}

 $\parallel l_i \parallel_1$ denotes the total number of samples of client $j$, the same as $\sum_{i=1}^Cn_j^i$.

\begin{figure}
    \centering
    \includegraphics[scale=0.15]{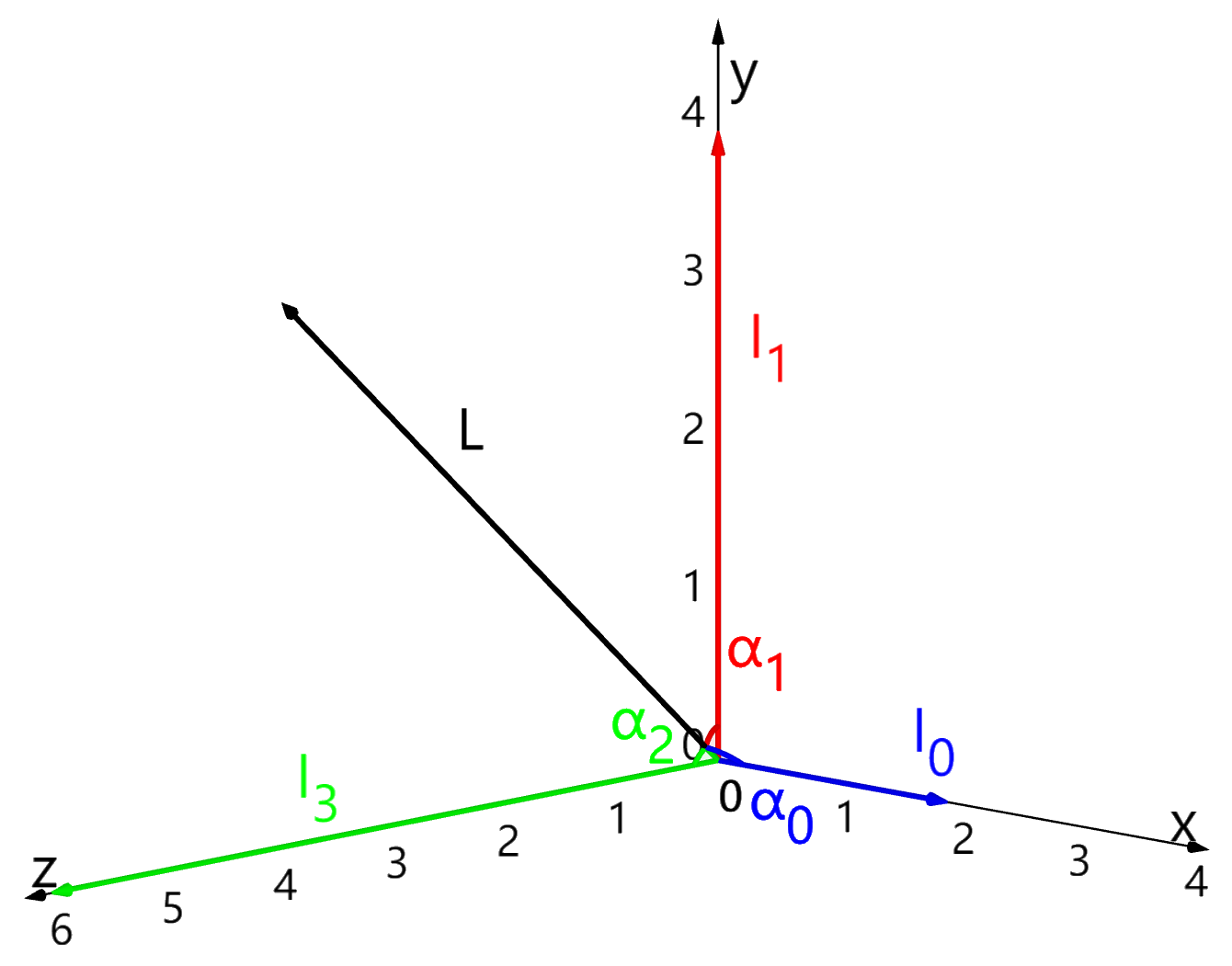}
    \caption{Weighted Cosine Similarity}
    \label{fig:wcs}
\end{figure}

For example, a federated learning network has 3 clients with label distribution vectors $l_1 = \left[2, 0, 0 \right]$, $l_2 = \left[0, 4, 0 \right]$, and $l_3 = \left[0, 0, 6 \right]$. As shown in Fig.~\ref{fig:wcs}, $L = \left[2,4,6\right]$. Follow Equation  \eqref{equ:wcs}, $L=\left[2,4,6\right]$. Let $\alpha_i$ be the angle between $L$ and $l_i$, then we have
\begin{equation*}
    WCS = \frac{2\cos{\alpha_1} + 4 \cos{\alpha_2} + 6 \cos{\alpha_3}}{2+4+6} = 0.62
\end{equation*}

Consider the extreme circumstance when the all distribution vectors $l_1, l_p$ are in same direction, we have $\alpha_i=0 (i\in\{1, \dots, P\})$ and $WCS=1$. When the label distribution vectors are completely in different directions $l_il_j=0(i \neq j)$ and same length, $WCS=\frac{1}{\sqrt{C}}$, which is also the minimum of $WCS$.

\section{Class Imbalanced Scenarios and Data Generation}
\label{sec:cis}

We use two metrics to define the class imbalance degree in Section III -- $MID$ and $WCS$. Based on the definitions, there exist 4 kinds of class imbalance scenarios.  
\begin{enumerate}
    \item $MID=0$, $WCS=1$ (scenario 1): the global data is strictly class balanced; all of the local label distribution vectors follow the same direction..
    \item $MID>0$, $WCS=1$ (scenario 2): the global data presents to be class imbalanced; all of the local label distribution vectors follow the same direction.
    \item $MID=0$, $WCS<1$ (scenario 3): the global data is strictly class balanced; the local label distribution vectors present discrepancy in directions.
    \item $MID>0$, $WCS<1$ (scenario 4): the global data presents to be class imbalanced; the local label distribution vectors present discrepancy in directions.
\end{enumerate}

When the local label distribution vectors presents discrepancy in directions, it means that some local datasets are class imbalanced. In the following experiments, we will discuss how the global model performs in these four scenarios, in particular the last three scenarios when either global or local data are class imbalanced. 

\subsection{Dataset Description and Preprocessing}
We select three popular datasets to simulate the three class imbalanced scenarios (scenarios 2-4) as identified above. 

\textbf{MNIST}~\cite{lecun1998gradient} is a handwriting digits dataset with 60000 train samples and 10000 test samples. Each data point consists of $28 \times 28$ gray pixels with a label range from 0 to 9. This dataset is split to 100 clients.

\begin{figure}
    \centering
    \input{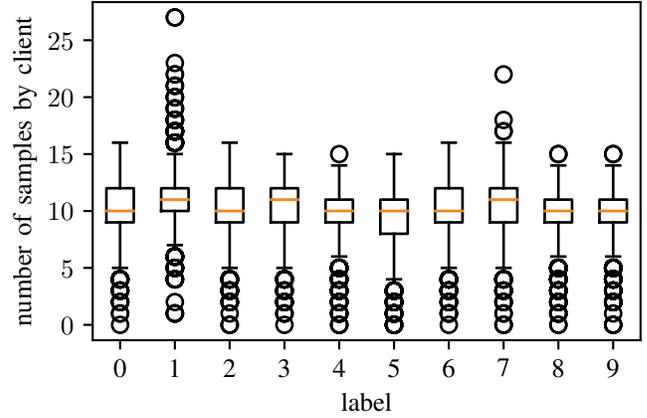}
    \caption{Nature of FEMNIST Dataset by Class}
    \label{fig:box}
\end{figure}

\textbf{FEMNIST}~\cite{caldas2018leaf} is a federated version of MNIST dataset with 341873 samples from 3383 writers which is clients in federated learning. With distinct writing styles, the data from different writers is non-iid. The box plot in Fig.~\ref{fig:box} shows the number of samples from different classes. As shown in this figure, the FEMNIST dataset is nearly globally class balanced, while local imbalance exists among clients due to the outliers. Each client in original FEMNIST dataset contains around 100 samples from 10 classes, which limits the experiments on extreme class imbalance when the local class imbalance ratio $\Gamma$ which is the ratio of majority and minority classes exceeds $10:1$. Therefore, 3383 clients of FEMNIST dataset are divided into 100 clients with a larger size local dataset that allows much more extreme local class imbalance degree.

\textbf{CIFAR10}~\cite{krizhevsky2009learning} dataset contains 50000 32x32 colour images in 10 classes. Each class has 5000 images. Similarly to MNIST, CIFAR10 was originally a centralized dataset. We split it to 100 clients randomly for federated learning. 

All the three datasets are globally balanced with 10 classes (labels 0 to 9). For MNIST and CIFAR10 datasets, each client contains the same number of samples for every class, i.e. locally balanced. FEMNIST dataset are nearly class balanced as each class contains around 10 samples per client. 

\subsection{Global Class Imbalance -- Scenario 2}
To investigate the impact of global class imbalance, we let all the local datasets have the same class imbalance degree.
To simulate more general case as in Focal Loss~\cite{Sarkar2020}, 4 classes are randomly selected as the minority classes (0, 1, 3, 6 classes).
Given a local imbalance ratio $\Gamma$, we randomly samples $\frac{1}{\Gamma}$ of data points of classes 0, 1, 3, 6 and keep all the data from classes 2, 4, 5, 7, 9. For the FEMNIST dataset, $\Gamma$ is set to $10:1$, $30:1$, and $100:1$ respectively. When $\Gamma=10:1$, every local dataset has only 3 data samples for classes 0, 1, 3, and 6. To make sure the local datasets contain samples from all the classes, $L$ is set to $10:1$, $20:1$, $60:1$ for MNIST dataset, and $10:1$, $20:1$, $50:1$ for CIFAR10 dataset.

\subsection{Local imbalance -- Scenario 3}
To simulate local class imbalance for MNIST and CIFAR10 datasets, We randomly select $S$ classes out of 10 for every client and evenly assign samples from the selected classes to that client. For example, CIFAR10 has 5000 training samples for each class. If $S$ is set to 2, every client contains 2 classes with 250 data points from each class. For FEMNIST, it is globally class balanced and contains class imbalanced local datasets as shown in Fig.~\ref{fig:box}.Instead of assigning all the samples to the clients, samples for classes that has not been selected are dropped from the balanced FEMNIST dataset.

\subsection{Global and Local Imbalance -- Scenario 4}
It is more common to have class imbalance at both local and global levels. We combine the data simulation steps from both Scenarios 2 and 3. Given an imbalance ratio $\Gamma$ bewteen majority and minority classes, for MNIST and CIFAR10, data samples from classes 0, 1, 3, 6 are sampled at a rate of $\Gamma$ while data from the other classes are all kept. Then the obtained imbalanced data is distributed to 100 clients with $S$ classes at each client. For FEMNIST, We randomly select $S$ classes for each client. For every client, if the selected class belong to one of the minority classes 0, 1, 3 and 6, the data is downsampled at the rate of $\frac{1}{\Gamma}$. The samples from non-selected classes are dropped.

\subsection{Summary of Generated Class Imbalanced Data}
In summary, we have generated 10 datasets that covers all four class imbalanced scenarios, based on each of the MNIST, FEMNIST and CIFAR10 datasets. Table~\ref{tab:cid} summarizes all the cases. $\Gamma$ presents the global class imbalance. $MID$ and $WCS$, as defined in the previous section, show the global class imbalance degree and the mismatch between local and global imbalance.

\begin{table}[t]
\centering
\caption{Simulated Class Imbalanced Data for Scenarios 1-4}
\begin{tabular}{lcrrrrr}
Dataset                   & Scenario           &$\Gamma$& S  & LRID   & MID   & WCS  \\ \hline \hline
\multirow{10}{*}{MNIST}   & 1                  & 1:1   & 10 & 171    & 0     & 1    \\ \cline{2-7}
                          & \multirow{3}{*}{2} & 10:1  & 10 & 22650  & 0.13  & 1    \\
                          &                    & 20:1  & 10 & 27758  & 0.17  & 1    \\
                          &                    & 60:1  & 10 & 32458  & 0.2   & 1    \\ \cline{2-7}
                          & \multirow{3}{*}{3} & 1:1   & 5  & 171    & 0     & 0.71 \\
                          &                    & 1:1   & 2  & 171    & 0     & 0.45 \\
                          &                    & 1:1   & 1  & 171    & 0     & 0.32 \\ \cline{2-7}
                          & \multirow{3}{*}{4} & 10:1  & 2  & 22650  & 0.13  & 0.49 \\
                          &                    & 20:1  & 2  & 27758  & 0.17  & 0.49 \\
                          &                    & 60:1  & 2  & 32458  & 0.2   & 0.5  \\ \hline
\multirow{10}{*}{CIFAR10} & 1                  & 1:1   & 10 & 0      & 0     & 1    \\ \cline{2-7}
                          & \multirow{3}{*}{2} & 10:1  & 10 & 19352  & 0.13  & 1    \\
                          &                    & 20:1  & 10 & 24696  & 0.17  & 1    \\
                          &                    & 50:1  & 10 & 27123  & 0.19  & 1    \\ \cline{2-7}
                          & \multirow{3}{*}{3} & 1:1   & 5  & 0      & 0     & 0.71 \\
                          &                    & 1:1   & 2  & 0      & 0     & 0.45 \\
                          &                    & 1:1   & 1  & 0      & 0     & 0.32 \\ \cline{2-7}
                          & \multirow{3}{*}{4} & 10:1  & 2  & 19352  & 0.13  & 0.48 \\
                          &                    & 20:1  & 2  & 23647  & 0.17  & 0.5  \\
                          &                    & 50:1  & 2  & 27123  & 0.19  & 0.49 \\ \hline
\multirow{10}{*}{FEMNIST} & 1                  & 1:1   & 10 & 761    & 0     & 1    \\ \cline{2-7}
                          & \multirow{3}{*}{2} & 10:1  & 10 & 129292 & 0.13  & 1    \\
                          &                    & 30:1  & 10 & 171270 & 0.18  & 1    \\
                          &                    & 100:1 & 10 & 193238 & 0.21  & 1    \\ \cline{2-7}
                          & \multirow{3}{*}{3} & 1:1   & 5  & 392    & 0     & 0.71 \\
                          &                    & 1:1   & 2  & 147    & 0     & 0.45 \\
                          &                    & 1:1   & 1  & 98     & 0     & 0.32 \\ \cline{2-7}
                          & \multirow{3}{*}{4} & 10:1  & 2  & 26479  & 0.13  & 0.48 \\
                          &                    & 30:1  & 2  & 34253  & 0.18  & 0.5  \\
                          &                    & 100:1 & 2  & 38650  & 0.21  & 0.51 \\ \hline \\
\end{tabular}
\label{tab:cid}
\end{table}

From Table~\ref{tab:cid}, when the majority and minority classes are fixed, $MID$ follow the trend the global class imbalance ration $\Gamma$. $MID$ is more informative than $\Gamma$, especially When there are not only majority and minority classes but also other classes between them. Compared with $LRID$, $MID$ is more stable with varies size of datasets.

\begin{table}[t]
\centering
\caption{Global performance of Scenarios 1 (baseline) and 2}
\begin{tabular}{llll}
Dataset                  & MID  & Accuracy & F1     \\ \hline \hline
\multirow{4}{*}{MNIST}   & 0    & 0.9893   & 0.9892 \\
                         & 0.13 & 0.9778   & 0.9777 \\
                         & 0.17 & 0.9686   & 0.9682 \\
                         & 0.2  & 0.9350   & 0.9326 \\ \hline
\multirow{4}{*}{CIFAR10} & 0    & 0.6254   & 0.6224 \\
                         & 0.13 & 0.4781   & 0.4247 \\
                         & 0.17 & 0.4389   & 0.3491 \\
                         & 0.19 & 0.4274   & 0.3264 \\ \hline
\multirow{4}{*}{FEMNIST} & 0    & 0.9916   & 0.9916 \\
                         & 0.13 & 0.9850   & 0.9849 \\
                         & 0.18 & 0.9772   & 0.9769 \\
                         & 0.21 & 0.9539   & 0.9532 \\ \hline \\
\end{tabular}
\label{tab:s2}
\end{table}

\begin{table}[t]
\centering
\caption{Global performance of the $2^{nd}$ case in Scenario 3 (baseline) and all cases in Scenario 4}
\begin{tabular}{lllll}
Dataset                  & MID  & WCS  & Accuracy & F1     \\  \hline  \hline
\multirow{4}{*}{MNIST}   & 0    & 0.45 & 0.9704   & 0.9703 \\
                         & 0.13 & 0.49 & 0.9312   & 0.9300 \\
                         & 0.17 & 0.49 & 0.9241   & 0.9227 \\
                         & 0.2  & 0.5  & 0.8598   & 0.8532 \\  \hline
\multirow{4}{*}{CIFAR10} & 0    & 0.45 & 0.4164   & 0.3880 \\
                         & 0.13 & 0.48 & 0.3315   & 0.2673 \\
                         & 0.17 & 0.5  & 0.3083   & 0.2268 \\
                         & 0.19 & 0.49 & 0.3066   & 0.2223 \\  \hline
\multirow{4}{*}{FEMNIST} & 0    & 0.45 & 0.8380   & 0.8323 \\
                         & 0.13 & 0.48 & 0.8247   & 0.8209 \\
                         & 0.18 & 0.5  & 0.6635   & 0.6337 \\
                         & 0.21 & 0.51 & 0.5375   & 0.4661 \\  \hline \\
\end{tabular}
\label{tab:s4}
\end{table}

\begin{figure*}[ht]
    \centering
    \input{fig/S2.pgf}
    \caption{Validation F1-Score of Scenarios 1 (baseline) and 2}
    \label{fig:s2}
\end{figure*}

\begin{figure*}[ht]
    \centering
    \input{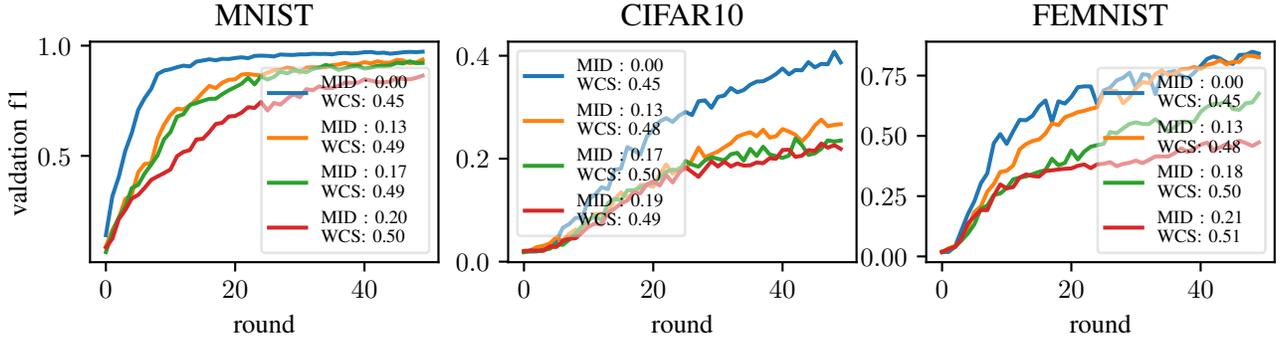}
    \caption{Validation F1-Score of the $2^{nd}$ case in scenario 3 (baseline) and all cases in Scenario 4}
    \label{fig:s4}
\end{figure*}

\begin{figure*}[ht]
    \centering
    \input{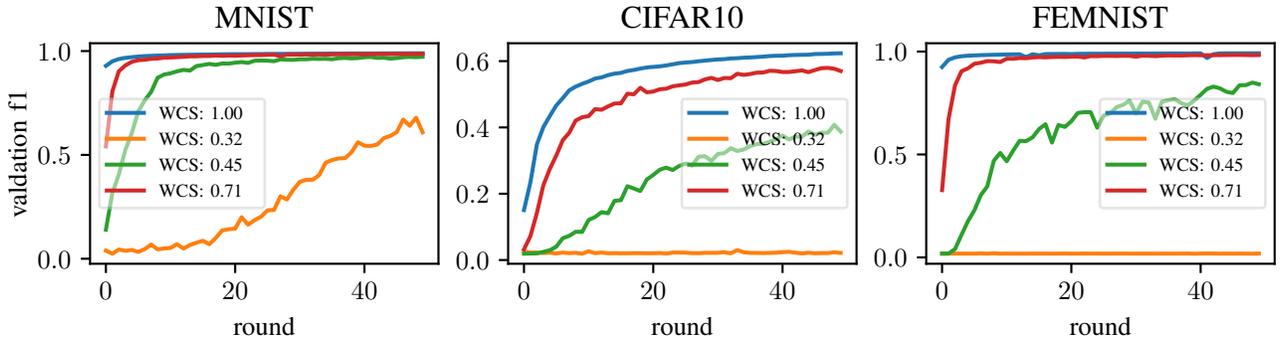}
    \caption{Validation F1-Score of Scenarios 1 (baseline) and 3}
    \label{fig:s3}
\end{figure*}

\section{Experiments}
\label{sec:exp}

We adopt FedAvg~\cite{BrendanMcMahan2017} algorithm to train a convolution neural networks (CNN) as the global model for each dataset. As MNIST, FEMNIST and CIFAR10 are all image datasets, we use the same CNN structure setting for all cases -- two convolutional layers followed by two dense layers.
At each round of training, 10 out of 100 clients are randomly selected to participate~\cite{Sarkar2020}. SGD is used in the local training as the optimizer with local learning rate equal to $0.1$. The server's learning rate is set to $1.0$ following the recommended value of TensorFlow Federated framework. Each global model is trained for 50 iterations with 5 local epochs every iteration at a batch size of 128. All three datasets can converge with this setting in scenario 1. In our experiment, the whole training process is repeated 15 times to get an average result.
As a class imbalance problem, we add F1-Score as an implement metrics to the overall accuracy to measure the model performance.

\subsection{Impact of Global Class Imbalance Degree}

Table~\ref{tab:s2} and Fig.~\ref{fig:s2} compare the performance of the global model in scenario 1 (baseline case) and scenario 2, where $WCS = 1$ and $MID$ varies between [0, 0.21]. Table~\ref{tab:s2} shows the final predictive accuracy and F1-Score. Fig.~\ref{fig:s2} shows the F1-Score curves along with training. The accuracy curves are very similar to the F1-Score ones, so they are omitted from the figure for the space reason. The results tell us how the global class imbalance degree impacts the global performance. We can observe a decrease of accuracy and F1-Score with the increase of global class imbalance degree $MID$ on all three datasets. Table~\ref{tab:s4} shows the accuracy and F1-Score of the global model in the second case from scenario 3 (baseline case) and all cases from scenario  4 with $S=2$, where $WCS \approx 0.5$ and $MID$ varies between [0, 0.21]. The corresponding F1-Score curves are presented in Fig.~\ref{fig:s4}. They also show that a larger global class imbalance degree reduces the global performance significantly.

\subsection{Impact of Local Class Imbalance Difference}

The difference of local class imbalance is a feature in federated learning that distinguishes itself from class imbalance problems in centralized machine learning. Table~\ref{tab:s3} shows the accuracy and F1-Score of the global model in scenario 1 (baseline case) and scenario 3, where $MID$ remains 0 and $WCS$ varies between [0.32, 1]. The corresponding F1-Score curves are presented in Fig.~\ref{fig:s3}. We observer that a smaller similarity between local class imbalance results in degradation of the global model performance. Besides, the decrease of $WCS$ results in a larger fluctuation on the performance curves especially in CIFAR10 and FEMNIST datasets as shown in Fig.~\ref{fig:s3} and Fig.~\ref{fig:s4} in comparison with Fig.~\ref{fig:s2}. As a result, the convergence of the global model is significantly slowed down. In scenario 3 when $WCS=0.32$ for CIFAR10 and FEMNIST (in the middle and right plots of Fig.~\ref{fig:s3}), the global model cannot even converge.

When comparing Scenario 4 (where $MID>0$ and $WCS<1$) with Scenario 2 (where $WCS=1$), we can see that a large difference of local class imbalance not only causes the degradation of the global model performance, but also introduces performance fluctuation and slows down the convergence of the global model. In short, by reducing $WCS$ (i.e. a larger difference), the global model becomes more difficult to converge and suffers worse prediction accuracy.


\begin{table}[t]
\centering
\caption{Global performance of Scenarios 1 (baseline) and 3}
\begin{tabular}{llllll}
Dataset                  & WCS  & Accuracy & F1     \\ \hline \hline
\multirow{4}{*}{MNIST}   & 1    & 0.9893   & 0.9892 \\
                         & 0.71 & 0.9863   & 0.9862 \\
                         & 0.45 & 0.9704   & 0.9703 \\
                         & 0.32 & 0.6718   & 0.6416 \\ \hline
\multirow{4}{*}{CIFAR10} & 1    & 0.6254   & 0.6224 \\
                         & 0.71 & 0.5830   & 0.5757 \\
                         & 0.45 & 0.4164   & 0.3880 \\
                         & 0.32 & 0.1026   & 0.0214 \\ \hline
\multirow{4}{*}{FEMNIST} & 1    & 0.9916   & 0.9916 \\
                         & 0.71 & 0.9833   & 0.9831 \\
                         & 0.45 & 0.8380   & 0.8323 \\
                         & 0.32 & 0.1012   & 0.0184 \\  \hline \\
\end{tabular}
\label{tab:s3}
\end{table}

\section{Conclusions}
\label{sec:con}

This paper investigates the impact of class imbalance in federated learning. We focus on three research questions: Q1. define class imbalance in federated learning. Q2. explore the impact of global class imbalance on the global model. Q3. explore the impact of imbalance differences between local clients on the global model.

For Q1, we proposed two new metrics -- MID and WCS. MID measures the global class imbalance degree. It improves the traditional Imbalance Ratio and LRID, which is suitable to multi-class data and is insensitive to the size of datasets. WCS is specifically designed for federated learning that measures the class imbalance differences among local clients and considers the contributions of local datasets. Based on MID and WCS, we looked into 4 class imbalanced scenarios to answer Q2 and Q3. For Q2, we found that a larger MID leads to more significant degradation of the global performance in terms of prediction accuracy and F1-Score. For Q3, we showed that a large difference of class imbalance degree among local datasets not only reduces the global performance, but also slows down the convergence by introducing fluctuation in optimization.

This work suggests that 
1) class imbalance in federated learning should be studied separately considering MID and WCS,
2) global class imbalance should be studied and tackled appropriately in federated learning for better global model performance, and 
3) the differences of local class imbalance should also be treated seriously that could affect the global performance and convergence speed.

\bibliographystyle{IEEEtran}
\bibliography{reference}

\end{document}